\documentclass[letterpaper, 10 pt, conference]{bib/ieeeconf}
\IEEEoverridecommandlockouts                              
\usepackage[normalem]{ulem}	                        
\usepackage[table,usenames,dvipsnames]{xcolor}      
\usepackage{extarrows}                              
\let\labelindent\relax
\usepackage{enumitem}                               

\usepackage{amsmath,amsthm,amssymb,amsfonts,dsfont} 
\usepackage{algorithm,algorithmicx,listings}        
\usepackage[noend]{algpseudocode}			        
\usepackage{graphicx}
\usepackage{subfigure}
\usepackage{tabularx}
\usepackage{amsmath}
\usepackage{multirow,multicol, array}
\usepackage[font={small}]{caption}   
\let\appendices\relax
\usepackage[titletoc,title]{appendix}

\usepackage[colorinlistoftodos,prependcaption,textwidth=1.5cm,textsize=tiny]{todonotes}
\newcommand{\naComment}[1]{}

\usepackage[breaklinks=true, colorlinks, bookmarks=true, citecolor=Black, urlcolor=Violet,linkcolor=Black]{hyperref}
\usepackage{booktabs}
\usepackage{cite}

\def\argmin{\mathop{\arg\min}\limits}	%
\newcommand{\prl}[1]{\mathopen{}\left(#1\right)\mathclose{}}

\newcommand{\T}{\mathsf{T}}

\newtheorem*{assumption*}{Assumption}

\newtheorem*{remark*}{Remark}
\newtheorem*{problem*}{Problem}
\newtheorem{problem}{Problem}

\title{\LARGE \bf
Search-based Motion Planning for Aggressive Flight in SE(3)}
\author{Sikang Liu$^{1}$,  Kartik Mohta$^{1}$, Nikolay Atanasov$^{2}$ and Vijay Kumar$^{1}$
\thanks{This work is supported by ARL \# W911NF-08-2-0004, DARPA \# HR001151626/HR0011516850, ARO \# W911NF-13-1-0350, and ONR \# N00014-07-1-0829.} 
\thanks{$^{1}$S. Liu, K. Mohta, and V. Kumar are with the GRASP Laboratory, University of Pennsylvania, USA 
 {\tt\small sikang@seas.upenn.edu}}%
\thanks{$^{2}$N. Atanasov is with the Electrical and Computer Engineering department, UC San Diego, USA
 {\tt\small natanasov@ucsd.edu}}%
}

\begin{document}
\maketitle
\begin{abstract}
Quadrotors with large thrust-to-weight ratios are able to track aggressive trajectories with sharp turns and high accelerations. In this work, we develop a search-based trajectory planning approach that exploits the quadrotor maneuverability to generate sequences of motion primitives in cluttered environments. We model the quadrotor body as an ellipsoid and compute its flight attitude along trajectories in order to check for collisions against obstacles. The ellipsoid model allows the quadrotor to pass through gaps that are smaller than its diameter with non-zero pitch or roll angles. Without any prior information about the location of gaps and associated attitude constraints, our algorithm is able to find a safe and optimal trajectory that guides the robot to its goal as fast as possible. To accelerate planning, we first perform a lower dimensional search and use it as a heuristic to guide the generation of a final dynamically feasible trajectory. We analyze critical discretization parameters of motion primitive planning and demonstrate the feasibility of the generated trajectories in various simulations and real-world experiments.
\end{abstract}

\section{Introduction}
\label{sec:intro}

\begin{figure}[t]
\centering
    \includegraphics[width=0.9\linewidth]{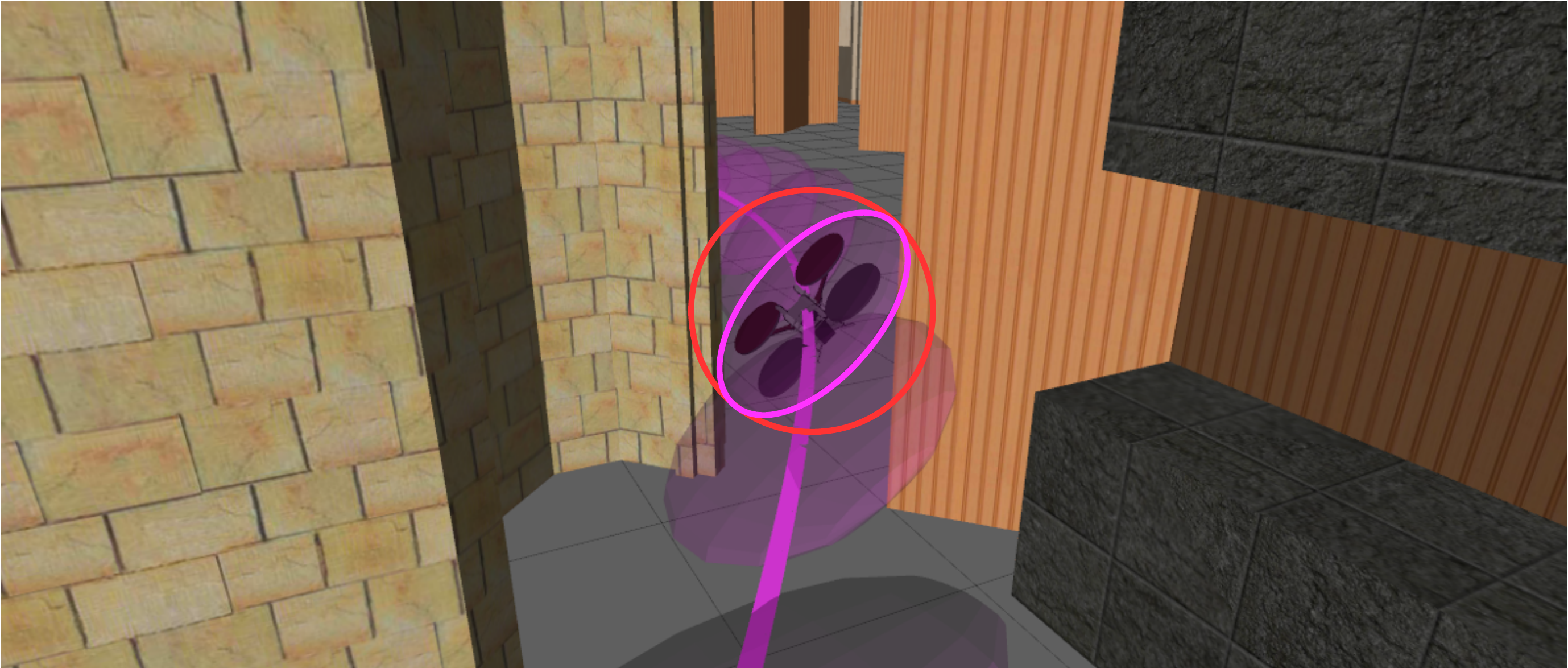}
     \caption{By taking the shape and dynamics of a quadrotor into account, our planner is able to generate a trajectory that allows the quadrotor to pass through a door, narrower than robot's diameter. In contrast, existing methods that model the quadrotor as a sphere (red circle) would not be able to find a feasible path in this environment.\label{fig: intro}}
\end{figure}

Motion planning, the problem of generating dynamically feasible trajectories that avoid obstacles in unstructured environments, for Micro Aerial Vehicles (MAVs), especially quadrotors, has attracted significant attention recently~\cite{liu2016high,richter2016polynomial,eth_planning_2016,deitsICRA2015}. When the MAV attitude and dynamics are taken into account, the problem is challenging because there are no simple geometric conditions for identifying collision-free configurations~\cite{canny_donald_reif_xavier_SFCS1988}. Existing planning approaches usually model the MAV as a sphere or prism, which allows obtaining a simple configuration space (C-space) by inflating the obtacles with the robot size. As a result, the robot can be treated as a single point in C-space and the collision-checking even for trajectories that take dynamics into account is simplified. Even though this spherical model assumption is widely used in motion planing, it is very conservative since it invalidates many trajectories whose feasbility depends on the robot attitude (Fig.~\ref{fig: intro}). Several prior works have demonstrated aggressive maneuvers for quadrotors that pass through narrow gaps~\cite{falanga2016aggressive,giuseppe_2017,Hirata_2014} but, instead of solving the planning problem, those works focus on trajectory generation with given attitude constraints. Those constraints are often hand-picked beforehand or obtained using gap detection algorithms which only works for specific cases.

We are interested in designing a planner that considers the robot's actual shape and dynamics in order to obtain aggressive trajectories in cluttered environments. Since quadrotors are under-actuated systems, they cannot translate and rotate independently. Thus, planners for fully-actuated system like spacecraft~\cite{how_2005,mike_2016} or omni-directional aerial vehicle~\cite{omni_2016} are not suitable for quadrotors. This paper builds on our previous search-based trajectory planning approach~\cite{liu_iros_2017} that utilizes motion primitives to discretize the control space and obtain a dynamically feasible resolution-complete (i.e., optimal in the discretized space) trajectory in cluttered environments. We extend our previous work by explicitly computing the robot attitude along the motion primitives and using it to enforce collision constraints. Furthermore, to reduce computation time for searching in high-dimensional (velocity, acceleration, jerk, etc.) space, we propose a novel hierarchical planning process that refines a dynamically feasible trajectory from a prior trajectory in lower dimensional space. The paper makes the following contributions:
\begin{enumerate}
\item A graph search algorithm that uses motion primitives to take attitude constraints into account and compute a dynamically feasible resolution-complete trajectory for a quadrotor is developed.
\item A hierarchical refinement process that uses prior lower-dimensional trajectories as heuristics to accelerate planning in higher dimensions is proposed.
\item The effect of motion primitive discretization parameters on the computation time, smoothness, and optimality of the generated trajectories is analyzed.
\end{enumerate}

The code used in this work is open-sourced at \url{https://github.com/sikang/mpl_ros}. Users can easily test our planner and benchmark the performance against other planning algorithms.


\section{Related Works}
\label{sec: related}
Trajectories for MAVs or, more generally, differentially flat systems~\cite{mellingerICRA2011} are usually represented as piecewise polynomials in time since their derivatives can be used to obtain explicit expressions for the system states and control inputs~\cite{van1997real}. When collision avoidance is taken into account, more constraints need to be added to the problem formulation to guarantee safety either through anchoring waypoints as in~\cite{richter2016polynomial, mellingerICRA2011} or building a safe flight corridor as in~\cite{deitsICRA2015,liu2017plan,chen2016online}. These approaches require planning in a C-space in which the robot's attitude does not affect collision checking. As described in the previous section, conservative symmetrical approximations of the robot body may ignore trajectories whose feasibility depends on the robot attitude. Hence, planning in SE(3) is necessary in order to obtain agile trajectories in cluttered environments. Planning with 6 degrees of freedom has been addressed in several works~\cite{ichnowski2015fast,ratliff2009chomp} via sampling techniques but these do not translate immediately to our problem, where the rotation and translation are coupled and a smooth, deterministic trajectory is desired. Methods based on motion primitives are a promising approach for planning dynamically feasible and collision-free trajectories. For example, lattice search with pre-defined primitives~\cite{pivtoraiko2009differentially,macallister2013path} may be used to plan trajecotires for non-circular robots in obstacle cluttered environments. In our previous work~\cite{liu_iros_2017}, we developed a global planning approach for quadrotors based on lattice search by using motion primitives generated via optimal control~\cite{mueller2015}. In this work, we extend~\cite{liu_iros_2017} to account for attitude constraints by explicitly computing the robot attitude along the motion primitives based on the desired acceleration and gravity.

While randomized sampling approaches have been effective at solving very high dimensional planning problems, they take a long time to converge to an optimal solution~\cite{karamanIJRR2011} and intermediate solution quality might be unpredictable. Hence, randomized approaches are not suitable for fast navigation in unknown environments where frequent, predictable re-planning is necessary. Traditional graph search techniques are considered inefficient in high dimensional spaces but appropriate heuristic design~\cite{hansen2007anytime,ara_star,aine2016multi} may accelerate their speed. Using weighted heuristics, however, produces sub-optimal solutions and does not always reduce planning time~\cite{wilt2012does}. An interesting, alternative idea for accelerating motion planning is based on adaptive dimensionality~\cite{gochev2011path}, which exploits preliminary search results in lower dimensions to accelerate the planning process in high dimensions. In~\cite{liu_iros_2017}, we used a trajectory refinement step that obtains a smooth (higher dimensional) trajectory from a trajectory planned in a lower dimensional space even though this refinement step can potentially lead to unsafe and infeasible trajectories. In this work, we use a hierarchical planning procedure---plan a trajectory in low dimensional space and use it as a heuristic to guide the search in high dimensional space---to replace the refinement step, while guaranteeing dynamical feasibility, safety, and resolution completeness.

\section{Motion Planning with Attitude Constraints}
\label{sec: approach}
In this section, we introduce our trajectory planning framework based on motion primitives. While our previous work~\cite{liu_iros_2017} guarantees safety, dyanmical feasibility and optimality, it assumes a spherical robot body. Here, we introduce a way to account for the robot attitude during planning based on the desired acceleration and gravity. Since the quadrotor yaw is decoupled and does not affect system dyanmics, we assume it remains constant during planning.

\subsection{System Dynamics in Planning}
Before introducing the planning approach, we inspect the relation between polynomial trajectories and system dynamics. The position $\mathbf{x} = [x, y, z]^\T$ in $\mathbb{R}^3$ of the quadrotor can be defined as a differentially flat output as described in~\cite{mellingerICRA2011}. The associated velocity $\mathbf{v}$, acceleration $\mathbf{a}$ and jerk $\mathbf{j}$ can be obtained by taking derivatives with respect to time as $\dot{\mathbf{x}}, \ddot{\mathbf{x}}, \dddot{\mathbf{x}}$ respectively. The desired trajectory for the geometric SE(3) controller as described in~\cite{lee2010geometric} can be written as $\Phi(t) = [\mathbf{x}_{d}^\T, \mathbf{v}_{d}^\T, \mathbf{a}_{d}^\T,\mathbf{j}_d^\T]^\T$. According to~\cite{hehn2011quadrocopter}, we assume the force and angular velocity are our control inputs to the quadrotor.  Ignoring feedback control errors, the desired mass-normalized force in the inertial frame can be obtained as
\begin{equation}\label{eq: force}
\mathbf{f}_d =  \mathbf{a}_{d} + g\mathbf{z}_w
\end{equation}
where $g$ is the gravitational acceleration and $\mathbf{z}_w = [0, 0, 1]^\T$ is the z-axis of the inertial world frame. 
Similar to~\cite{lee2010geometric}, given a specific yaw $\psi$, the desired orientation in SO(3) can be written as $\mathbf{R}_d = [\mathbf{r}_1, \mathbf{r}_2, \mathbf{r}_3]$ where
\begin{equation}\label{eq: rotation}
\mathbf{r}_3 = \mathbf{f}_d / \|\mathbf{f}_d\|, \quad \mathbf{r}_1 = \frac{\mathbf{r}_{2c} \times \mathbf{r}_{3}}{\|\mathbf{r}_{2c} \times \mathbf{r}_{3}\|}, \quad \mathbf{r}_2 = \mathbf{r}_3 \times \mathbf{r}_1
\end{equation}
and
\begin{equation}
\mathbf{r}_{2c} = [-\sin{\psi}, \cos{\psi}, 0]^\T
\end{equation} 
which is assumed to be not parallel to $\mathbf{r}_3$. The associated angular velocity in the inertial frame, $\dot{\mathbf{R}}_d = [\dot{\mathbf{r}}_1, \dot{\mathbf{r}}_2, \dot{\mathbf{r}}_3]$, can be calculated as
\begin{align}
\dot{\mathbf{r}}_3 &= \mathbf{r}_3\times \frac{\dot{\mathbf{f}}_d}{\|\mathbf{f}_d\|}\times \mathbf{r}_3,\nonumber\\
\dot{\mathbf{r}}_1 &= \mathbf{r}_1\times \frac{\dot{\mathbf{r}}_{2c} \times \mathbf{r}_{3}+\mathbf{r}_{2c}\times \dot{\mathbf{r}}_3}{\|\mathbf{r}_{2c}\times\mathbf{r}_3\|}\times\mathbf{r}_1,\\
\dot{\mathbf{r}}_2 &= \dot{\mathbf{r}}_3 \times \mathbf{r}_1 + \mathbf{r}_3\times\dot{\mathbf{r}}_1\nonumber
\end{align}
where 
\begin{equation}
\dot{\mathbf{r}}_{2c} = [-\cos{\psi}, -\sin{\psi}, 0]^\T\dot{\psi}, \;\dot{\mathbf{f}_d} = \mathbf{j}^\T_d
\end{equation}

Therefore, the desired angular velocity $\mathbf{w}_d$ in body frame is obtained as:
\begin{equation}
 [\mathbf{w}_d]_{\times}= \mathbf{R}^\T_d\dot{\mathbf{R}}_d
\end{equation}

Once the desired force $\mathbf{f}_d$, orientation $\mathbf{R}_d$ and angular velocity $\mathbf{w}_d$ are defined, it is straightforward to compute the desired control inputs for the quadrotor system. Notice that: 1) orientation is algebraically related to the desired acceleration and gravity and 2) angular velocity is algebraically related to the desired jerk.

\subsection{Search-based Planning using Motion Primitives}\label{sec: plan}
As mentioned in the previous section, the desired trajectory can be defined as 
\begin{equation}
\Phi(t) := [\mathbf{x}^\T, \dot{\mathbf{x}}^\T, \ddot{\mathbf{x}}^\T, \dddot{\mathbf{x}}^\T]^\T = [\mathbf{x}^\T, \mathbf{v}^\T, \mathbf{a}^\T,\mathbf{j}^\T]^\T
\end{equation}
and each component of $\Phi(t)$ can be represented by a polynomial parameterized in time $t$. Position can be defined as
\begin{equation}\label{eq: poly}
\mathbf{x}(t) :=  \sum_{k=0}^{K} d_k\frac{t^k}{k!} = d_K \frac{t^K}{K!}+ \ldots + d_1 t + d_0
\end{equation}
where $d_k \in \mathbb{R}^3$ are the coefficients. The corresponding velocity, acceleration and jerk can be obtained by taking the derivative of~\eqref{eq: poly}. A polynomial trajectory from one state to the other within a specified time duration is called a \textit{motion primitive}. Our approach uses primitives generated as the solutions to an optimal control problem~\cite{liu_iros_2017} to build a graph from an initial state to a goal state and search for the optimal sequence of primitives. Technical details and proof of optimality can be found in our previous work~\cite{liu_iros_2017}. In this paper, we give the explicit solution for generating the optimal trajectory using jerk as the control input. 

We define the state 
\begin{equation}
\mathbf{s}(t) := [\mathbf{x}(t)^\T, \dot{\mathbf{x}}(t)^\T, \ddot{\mathbf{x}}(t)^\T]^\T = [\mathbf{p}^\T, \mathbf{v}^\T, \mathbf{a}^\T]^\T
\end{equation}
as a subset of the trajectory $\Phi(t)$ that excludes the jerk. From an initial state $\mathbf{s}_0 = [\mathbf{p}_0^\T, \mathbf{v}_0^\T, \mathbf{a}_0^\T]^\T$, we apply a constant jerk input $\mathbf{u}_m$ from a pre-defined control set $\mathcal{U}_M$ for a short duration $\tau > 0$. The resulting curve between $\mathbf{s}_0$ and the end state is a motion primitive such that for $t\in[0, \tau]$ the system state $\mathbf{s}(t)$ can be written as
\begin{equation}\label{eq:polynomial}
\mathbf{s}(t) = F(\mathbf{u}_m, \mathbf{s}_0, t) := 
\begin{bmatrix}
\mathbf{u}_m\frac{t^3}{6} + \mathbf{a}_0\frac{t^2}{2} + \mathbf{v}_0 t + \mathbf{p}_0\\
\mathbf{u}_m\frac{t^2}{2} + \mathbf{a}_0t + \mathbf{v}_0\\
\mathbf{u}_mt + \mathbf{a}_0
\end{bmatrix}
\end{equation}
It has been shown in~\cite{mueller2015} and~\cite{liu_iros_2017} that $F(\cdot)$ provides the minimum jerk trajectory between $\mathbf{s}_0$ and $\mathbf{s}(\tau)$.

\begin{figure}[htp]
  \centering
  \subfigure[$\tau = 0.5$,  $|\mathcal{U}_M| = 9$]{\includegraphics[width=0.45\columnwidth, trim=0 0 0 0, clip=true]{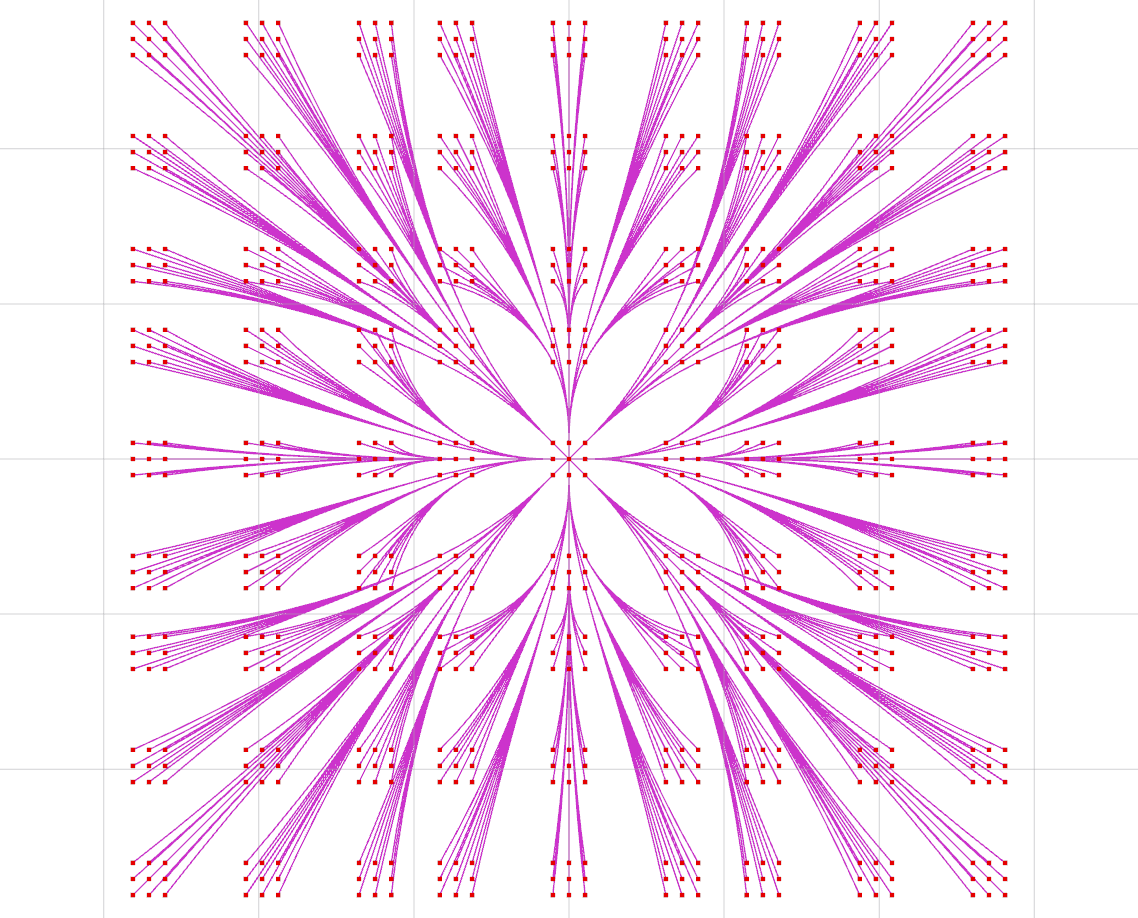}\label{fig: s1}}
  \subfigure[$\tau = 0.5$,  $|\mathcal{U}_M| = 25$]{\includegraphics[width=0.45\columnwidth, trim=0 0 0 0, clip=true]{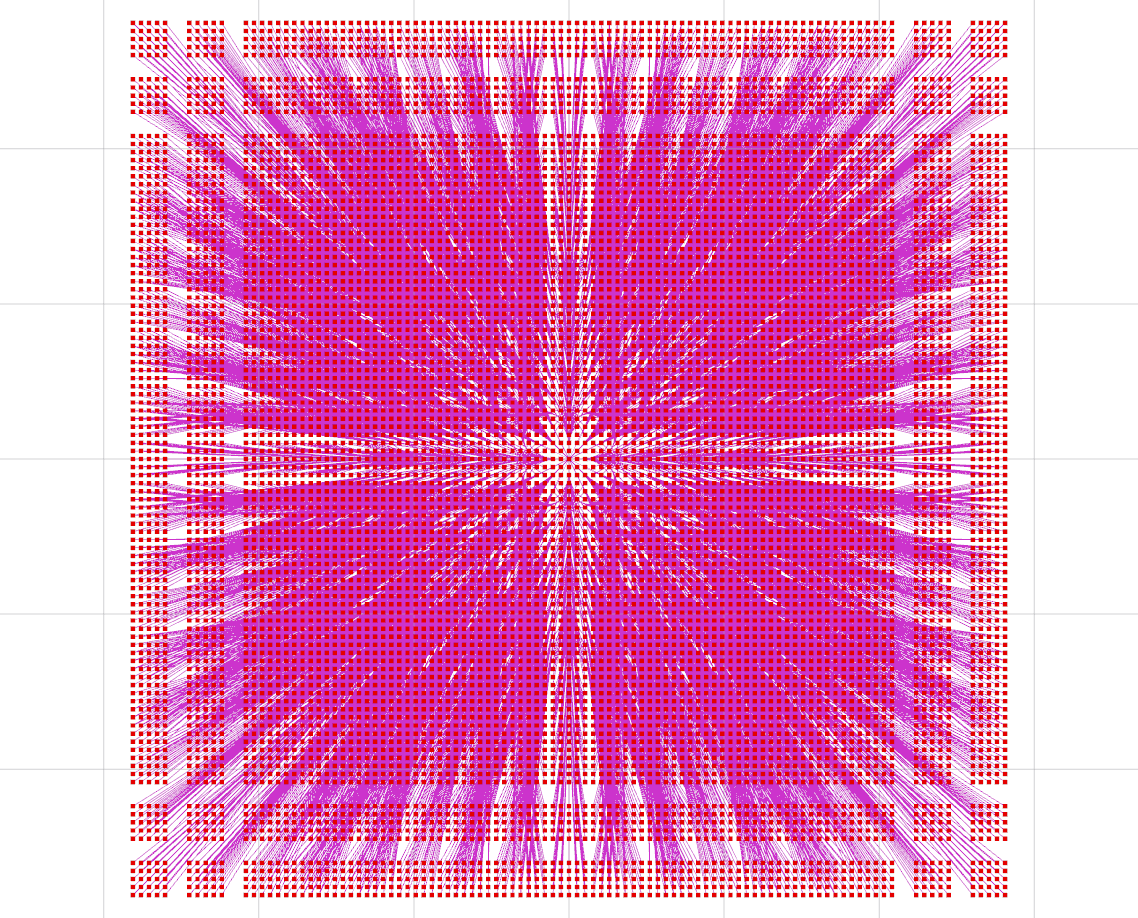}\label{fig: s2}}
   \caption{Graph $\mathcal{G}(\mathcal{S}, \mathcal{E})$ generated by applying BFS for a finite planning horizon over a set of motion primitives $\mathcal{U}_M$ with 9 elements (a) and 25 elements (b). Red dots represent states in $\mathcal{S}$ and magenta splines represent edges in $\mathcal{E}$.  \label{fig: graph}}
\end{figure}

The finite control input set $\mathcal{U}_M$ and duration $\tau$ define a graph $\mathcal{G}(\mathcal{S}, \mathcal{E})$, where $\mathcal{S}$ is the set of reachable states in $\mathbb{R}^{9}$ and $\mathcal{E}$ is the set of edges connecting those states. The states in $\mathcal{S}$ are generated by applying each element of $\mathcal{U}_M$ at each state iteratively, and each element in $\mathcal{E}$ is a primitive as defined in~\eqref{eq:polynomial}. A breadth-first-search (BFS) of a finite horizon leads to the graphs shown in Fig.~\ref{fig: graph}.

We are interested in finding a trajectory from $\mathbf{s}_0$ to $\mathbf{s}_g$ that is optimal in terms of total control effort $J$ and time $T$ taken to reach the goal. According to~\cite{liu_iros_2017}, a desired optimal trajectory is obtained as
\begin{equation}
    \label{eq:optimal}
    \begin{gathered}
      \Phi^*(t) = \argmin_{\Phi(t)}J+ \rho T = \argmin_{\Phi(t)}\int_0^T \|\mathbf{j}\|^2+ \rho T\\
      \begin{aligned}
        \text{s.t.}\;&\mathbf{s}_0 \leftarrow \Phi(0),\; \mathbf{s}_g \leftarrow \Phi(T)\\
      \end{aligned}
    \end{gathered}
  \end{equation}
where $\rho$ is the weight that decides the trade-off between effort and time.

For the primitive defined in~\eqref{eq:polynomial}, $J = \|\mathbf{u}_m\|^2\tau$ and $T = \tau$. Thus, the cost of a primitive of applying $\mathbf{u}_m$ from state $\mathbf{s}_n\in\mathcal{S}$ is defined as 
\begin{equation}
C(\mathbf{s}_n, \mathbf{u}_m) = C(\mathbf{u}_m) = (\|\mathbf{u}_m\|^2 + \rho)\tau
\end{equation}
 The cost of the individual primitive is independent of the current state and only depends on the set $\mathcal{U}_m$ and $\tau$. In addition, it can be shown by Pontryagin' minimum principle that~\eqref{eq:polynomial} is the optimal solution of~\eqref{eq:optimal}. Details of the proof can be found in~\cite{liu_iros_2017}. \naComment{This whole paragraph is unclear... Why is $\Phi$ the optimization variable in (11)? Where are the constraints in (11)? Where does the cost $J-\|u_m\|^2$ come from} Therefore, search for an optimal trajectory of~\eqref{eq:optimal} is equivalent to find the optimal solution to the following deterministic shortest path problem:
\begin{problem}\label{prob:1}
  Given an initial state $\mathbf{s}_0$, a goal region $\mathcal{X}^{goal}$, a free space $\mathcal{X}^{free}$ and motion primitives based on a finite set of control inputs $\mathcal{U}_M$ with duration $\tau > 0$, choose a sequence of control inputs $\mathbf{u}_{0:N-1}$ of length $N$ such that:  
  \begin{equation}
    \label{eq:problem}
    \begin{gathered}
      \min_{N, \mathbf{u}_{0:N-1}} \; \prl{\sum_{n=0}^{N-1}  \|\mathbf{u}_n\|^2 + \rho N}\tau\\
      \begin{aligned}
        \text{s.t.}\;&F_n(t) := F(\mathbf{u}_n, \mathbf{s}_n, t),\; \mathbf{u}_{n} \in \mathcal{U}_M\\
        \;&\mathbf{s}_{n+1} = F_n(\tau) = F_{n+1}(0), \;\mathbf{s}_{N} \in \mathcal{X}^{goal}\\
        &F_n(t) \subset \mathcal{X}^{free}
      \end{aligned}
    \end{gathered}
  \end{equation}
\end{problem}

We are able to solve this problem through a graph search algorithm like A*. The optimal trajectory $\Phi^*(t)$ can be recovered by applying the optimal control solution $\mathbf{u}^*_{0:N-1}$ with~\eqref{eq:polynomial} from the start $\mathbf{s}_0$ as
\begin{equation}
\Phi^*(t) \leftarrow [\mathbf{s}_0\overset{\mathbf{u}^*_0}{\longrightarrow} \mathbf{s}_1 \hdots \overset{\mathbf{u}^*_{N-1}}{\longrightarrow} \mathbf{s}_{N}]
\end{equation}

When planning dynamic trajectories, traditional distance-based heuristics are not effective since short-distance trajectories may require sudden changes in velocity, acceleration or orientation. Instead, we use a heuristic, proposed in~\cite{liu_iros_2017}, which is based on the solution of a \textit{Linear Quadratic Minimum Time} (LQMT) problem and takes trajectory smoothness into account. Given the current state $\mathbf{s}$ and the goal state $\mathbf{s}_g$, the LQMT solution provides an explicit formula for the $H(\mathbf{s}, \mathbf{s}_g)$ as described in Appendix~\ref{app:a}.



\subsection{Feasibility Checking}
When checking if a motion primitive is contained in the free space $\mathcal{X}^{free}$ in Problem~\ref{prob:1}, we need to consider both dynamical constraints that arise from system dynamics and geometric constraints due to physical obstacles. 

\subsubsection{Dynamically Feasible Primitives}
The dynamical constraints on a quadrotor system are the min/max thrust and torques that can be provided by the motors~\cite{mueller2015}. However, it is hard to examine the true specification for each quadrotor and apply correct non-linear constraints. In fact, it is reasonable to utilize the property of differential flatness and apply velocity, acceleration, and jerk constraints on each axis independently. This leads to componentwise inequalities of the form:
\begin{equation}\label{eq: ineq}
|\dot{\mathbf{x}}(t)| \preceq \bar{\mathbf{v}}_{max},\; |\ddot{\mathbf{x}}(t)| \preceq \bar{\mathbf{a}}_{max},\;|\dddot{\mathbf{x}}(t)| \preceq \bar{\mathbf{j}}_{max}
\end{equation}

Polynomial expressions for $\mathbf{\dot x}, \mathbf{\ddot x}, \mathbf{\dddot x}$ allow us to check~\eqref{eq: ineq} in \textit{closed-form} for each axis by finding the min/max value on time interval $[0, \tau]$. The latter is equivalent to finding the roots of the corresponding derivatives. Thus, we can guarantee that the planned trajectories always stay within the bounds $\bar{\mathbf{v}}_{max}, \bar{\mathbf{a}}_{max}, \bar{\mathbf{j}}_{max}$.

\subsubsection{Collision Free Primitives}
\label{sec: collision}
\begin{figure}[htp]
  \centering
  \subfigure{\includegraphics[width=0.9\columnwidth, trim=0 0 0 0, clip=true]{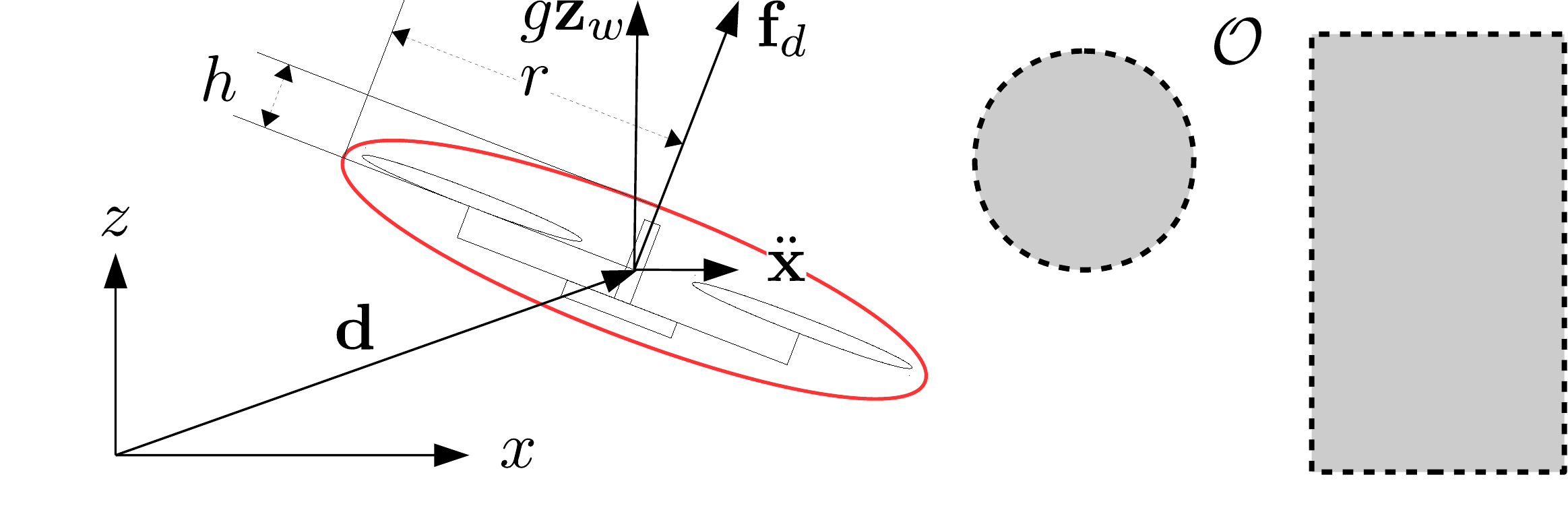}}
   \caption{A quadrotor can be modeled as an ellipsoid with radius $r$ and height $h$. Its position and attitude can be estimated from the desired trajectory. A point cloud $\mathcal{O}$ is used to represent obstacles. \label{fig:schematic}}
\end{figure}

Traditional collision checking is implemented in occupancy grid maps where the free and occupied spaces are discretized into cells. The robot is usually assumed to have a spherical shape. By inflating occupied cells with the radius of the robot, we are able to treat the robot as a single cell and check the occupancy of cells only along a trajectory. As mentioned in Section~\ref{sec:intro}, this process is too conservative and not suitable for planning agile trajectories in cluttered environments since it fails to take the actual robot shape and attitude into account. In this paper, we model the quadrotor as an ellipsoid $\xi$ in $\mathbb{R}^3$ with radius $r$ and height $h$ and the obstacle map as a point cloud $\mathcal{O}\subset \mathbb{R}^3$ (Fig.~\ref{fig:schematic}). Given a quadrotor state $\mathbf{s}$, its body configuration $\xi$ at $\mathbf{s}$ can be obtained as
\begin{equation}
\xi(\mathbf{s}) := \{\mathbf{p} = \mathbf{E}\tilde{\mathbf{p}}+\mathbf{d}\;|\;\|\tilde{\mathbf{p}}\|\leq 1\}
\end{equation}
where
\begin{equation}
\mathbf{d} = \mathbf{x}(t),\;
\mathbf{E} = \mathbf{R}\begin{bmatrix}
r & 0 & 0 \\
0 & r & 0 \\
0 & 0 & h
\end{bmatrix}\mathbf{R}^\T
\end{equation}
and the orientation $\mathbf{R}$ can be calculated from $\ddot{\mathbf{x}}(t)$ and gravity as shown in~\eqref{eq: rotation}.

Checking whether the quadrotor hits obstacles while following a trajectory is equivalent to checking if there is any obstacle inside the ellipsoid along the trajectory. In other words, we need to verify that the intersection between $\xi$ and the point cloud $\mathcal{O}$ is empty:
\begin{equation}
\mathcal{O} \cap \xi = \{\mathbf{o}\; |\; \|\mathbf{E}^{-1}(\mathbf{o} - \mathbf{d})\| \leq 1,\; \forall \mathbf{o}\in\mathcal{O}\}= \varnothing
\end{equation}
Instead of checking through every point in $\mathcal{O}$, it is more efficient to use \textit{KD-tree}~\cite{Rusu_ICRA2011_PCL} to crop a subset $\mathcal{O}_{r, \mathbf{d}}$ of $\mathcal{O}$ at first and then check the intersection between $\xi$ and obstacles inside $\mathcal{O}_{r, \mathbf{d}}$.  The subset $\mathcal{O}_{r, \mathbf{d}}$ is created by looking for neighbor points around $\mathbf{d}$ within radius $r$, assuming $r \geq h$. 

Since the contour of an ellipsoid following a primitive is not convex, we sample $I$ states in time along a primitive $F_n$ and consider the primitive $F_n$ collision-free if
\begin{equation}
\mathcal{O}\cap \xi(\mathbf{s}_{i, n}) = \varnothing, \; \forall i = \{0, 1, \hdots, I-1\}
\end{equation}
where $\mathbf{s}_{i, n}$ is the $i$-th sampled state on $F_n$.



In sum, the explicit formulation of the feasibility constraints $F_n(t) \subset \mathcal{X}^{free}$ in Problem~\ref{prob:1} is written as:
\begin{align}
 {F}_n(t) \preceq [\bar{\mathbf{v}}_{max}^\T, \bar{\mathbf{a}}_{max}^\T,\bar{\mathbf{j}}_{max}^\T]^\T\\
        \mathcal{O}\cap \xi(\mathbf{s}_{i, n}) = \varnothing, \; \forall i = \{0, 1, \hdots, I\} \nonumber
\end{align}

\section{Trajectory Refinement}
\label{sec: refine}
In the proposed planning approach, the dimension of the state space increases with increasing requirements on the continuity of the final trajectory. More precisely, if $\text{C}^2$ continuity is required for the final trajectory, jerk should be used as a control input and the state space of the associated second order system would be $\mathbb{R}^{9}$ (position, velocity acceleration). Generally, planning in higher dimensional spaces (e.g., snap input) requires more time and memory to explore and store states. In this section, we introduce a hierarchical approach to planning a feasible trajectory in high dimensional space by utilizing guidance from a trajectory planned in lower dimensional space. We show that the overall computation time of this hierarchical planning is shorter than the total time it takes to plan a optimal trajectory directly. Due to the fact that the final trajectory is calculated from a trajectory in lower dimensional space, similar to the refinement process in~\cite{liu_iros_2017}, we call this hierarchical planning process--trajectory refinement.

\subsection{Trajectories Planned in Different Control Spaces}
\begin{figure}[t]
  \centering
  \subfigure[$\Phi^1: T^1 = 32s, J^1 = 42, t^1= 2ms$]{\includegraphics[width=0.9\columnwidth, trim=0 0 0 0, clip=true]{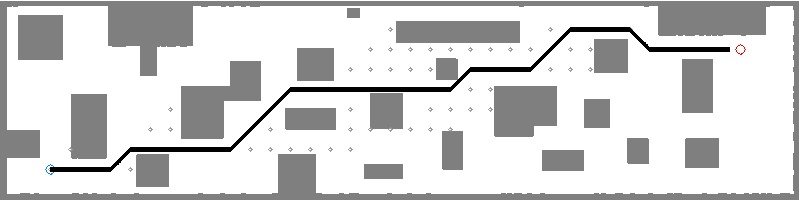}\label{fig: vel}}
  \subfigure[$\Phi^2: T^2 = 33s, J^2 = 2.25, t^2 = 60ms$]{\includegraphics[width=0.9\columnwidth, trim=0 0 0 0, clip=true]{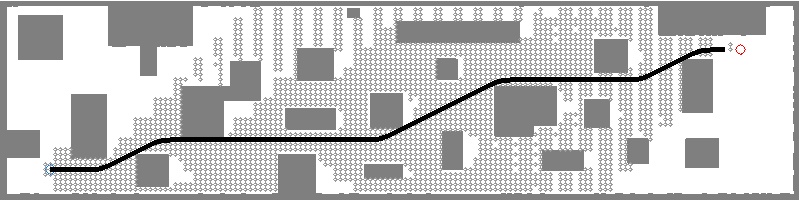}\label{fig: acc}}
  \subfigure[$\Phi^3: T^3 = 34s, J^3 = 3.75, t^3 = 1646ms$]{\includegraphics[width=0.9\columnwidth, trim=0 0 0 0, clip=true]{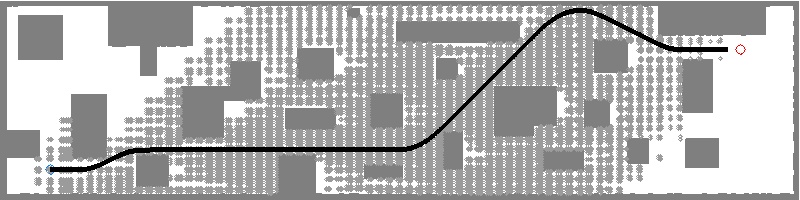}\label{fig: jrk}}
   \caption{Optimal trajectories planned using piecewise constant (a) velocity, (b) acceleration, (c) jerk from a start (blue dot) to a goal (red dot) state. Grey dots indicates explored states. \label{fig: trajs}}
\end{figure}

Denote the trajectories planned using velocity, acceleration or jerk inputs as $\Phi^{j}, j = 1,2,3$ respectively. Given the same start and goal, dynamics constraints and discretization, examples of the optimal trajectories in each case are plotted in Fig.~\ref{fig: trajs}, where the control effort $J^{j},j = {1, 2, 3}$ of the whole trajectory is measured as 
\begin{equation}\label{eq: J}
J^j = \int_0^{T}\|\mathbf{x}^{(j)}\|^2dt.
\end{equation}

Denote the execution and computation time of the trajectory as $T^j$ and $t^j, j=1,2,3$ accordingly. From the planning results in Fig.~\ref{fig: trajs}, two conclusions can be drawn with increasing $j$:
\begin{enumerate}
\item The execution time increases, i.e $T^1 < T^2 < T^3$;
\item The computation time increases, i.e $t^1 < t^2 < t^3$.
\end{enumerate}
Note that the computation time increases dramatically as $j$ increases. 

\subsection{Using Trajectories as Heuristics}

Denote the prior trajectory in lower dimensional space as $\Phi^{p}$. It is easy to obtain the set of a sequence of waypoints from start to goal as $\Phi^p \rightarrow [\mathbf{s}^{p}_0, \mathbf{s}^p_1,\hdots,\mathbf{s}^p_{N^p}]$, of which each element $\mathbf{s}^p_n$ is evaluated on $\Phi^p$ at the time $T_n = n\tau$. When searching for a trajectory in higher dimensional space ($q > p$) $\Phi^q:= [\mathbf{s}^{q}_0, \mathbf{s}^{q}_1,\hdots,\mathbf{s}^{q}_{N^q}]$, we propose to use the following heuristic function:
\begin{equation}\label{eq: H}
H(\mathbf{s}_n^q, \Phi^p) = H_1(\mathbf{s}_n^q, \mathbf{s}^p_n) + H_2(\mathbf{s}^p_n, \mathbf{s}^p_{N^p})
\end{equation}
The first term $H_1(\cdot)$ on the RHS of~\eqref{eq: H} is proposed in Appendix~\ref{app:a} where $\mathbf{s}_n^q$ is fully defined but $\mathbf{s}^p_n$ has undefined states. The second term $H_2(\cdot)$ is given directly as the cost from $\mathbf{s}^p_n$ to the goal by following $\Phi^p$, thus
\begin{equation}
H_2(\mathbf{s}^p_n, \mathbf{s}^p_{N^p}) = J^q(\mathbf{s}^p_n, \mathbf{s}^p_{N^p}) + \rho (T^p - T_n)
\end{equation}
where $T^p$ is execution time of $\Phi^p$ and $J^q(\mathbf{s}^p_n, \mathbf{s}^p_{N^p})$ is the control effort from $\mathbf{s}^p_n$ to $\mathbf{s}^p_{N^p}$ along $\Phi^p$ at order $q$ as~\eqref{eq: J}. This formulation is consistent with the cost function defined before in~\eqref{eq:optimal}. 
As the prior trajectory is in the lower dimensional space, $J^q$ for $\Phi^p$ is always zero. 
Thus $H_2$ is the execution time between $\mathbf{s}^p_n$ and $\mathbf{s}^p_{N^p}$:
\begin{equation}
H_2(\mathbf{s}^p_n, \mathbf{s}^p_{N^p}) = \rho(T^p - T_n)
\end{equation}

\begin{figure}[htp]
  \centering
  \subfigure[$x-y$ plot]{\includegraphics[width=0.45\columnwidth, trim=0 0 0 0cm, clip=true]{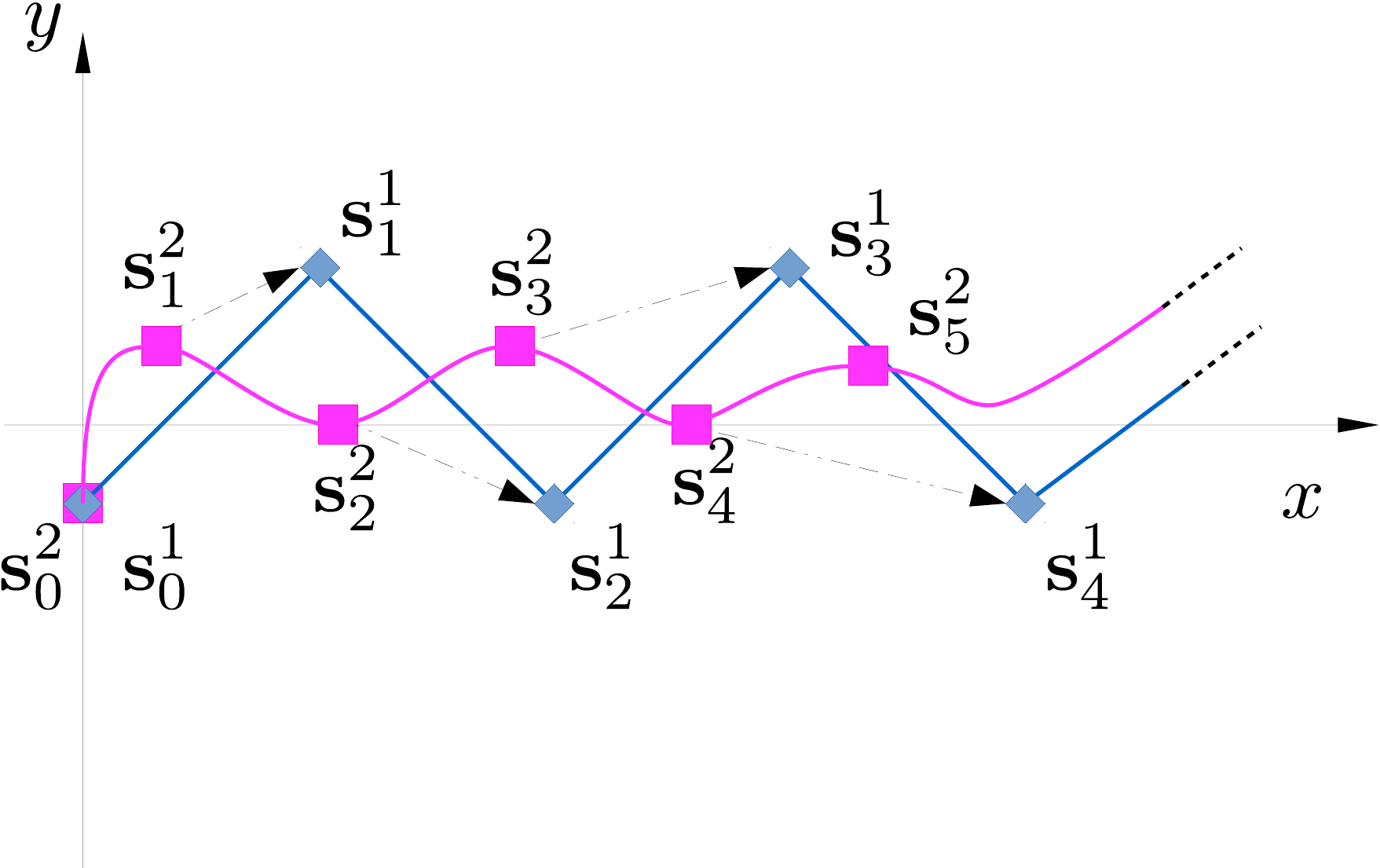}}
  \subfigure[$t-x$ plot]{\includegraphics[width=0.45\columnwidth, trim=0 0 0 0cm, clip=true]{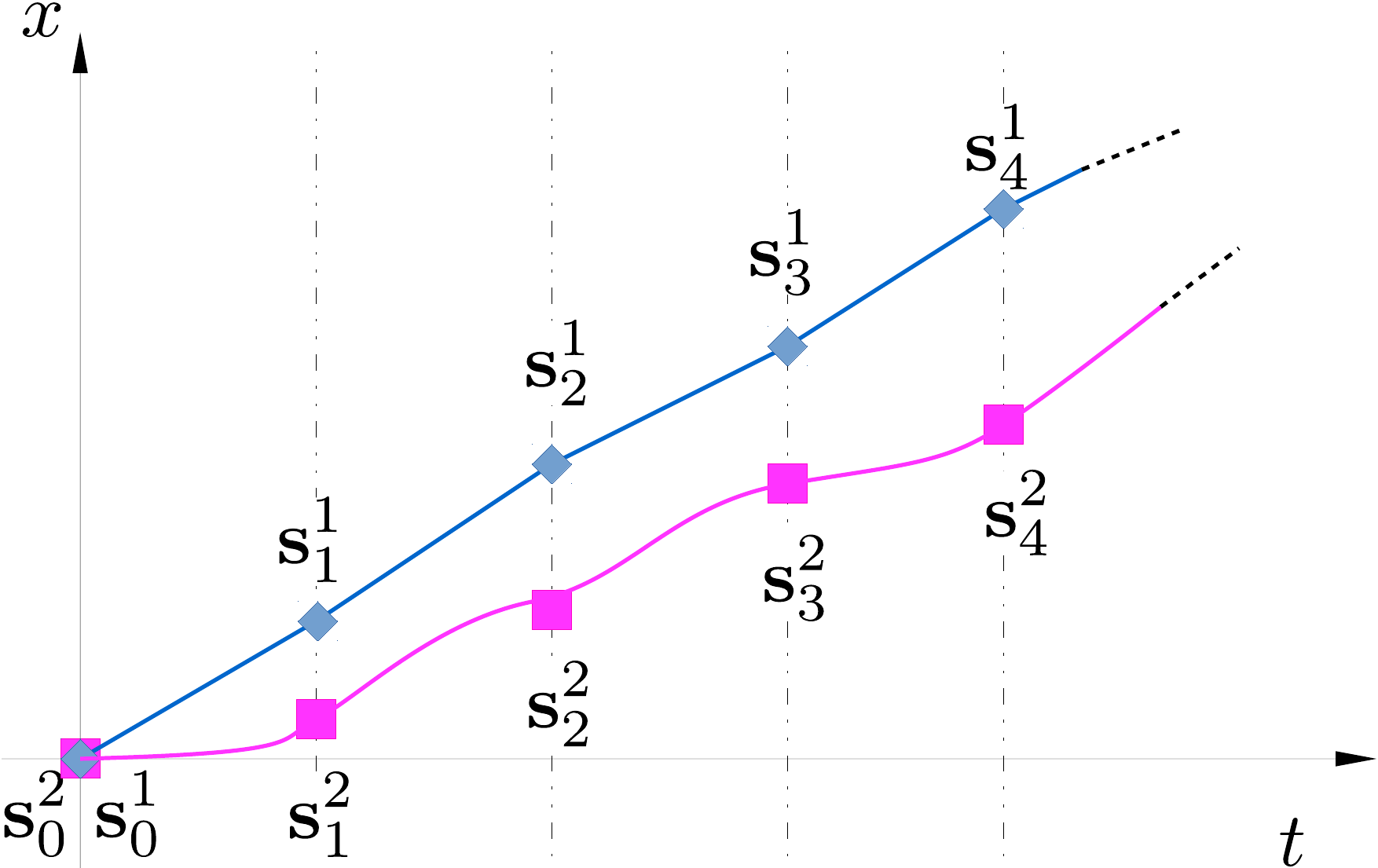}}
   \caption{Search $\Phi^2$ (magenta) using $\Phi^1$ (blue) as the heurisric. Left figure plots the trajectories in $x-y$ plane, the black arrows indicate the $H_1$. Right figure shows the corresponding $x$ position with respect to time $t$ along each trajectory, for states with the same subscript, they are at the same time $T_n$.\label{fig: hs}}
\end{figure}

Fig.~\ref{fig: hs} shows an example of applying~\eqref{eq: H} to search a trajectory $\Phi^2$ using acceleration with a prior trajectory $\Phi^1$ planned using velocity. Apparently, the new trajectory $\Phi^2$ will try to stick close to the prior trajectory $\Phi^1$ due to the effect of~\eqref{eq: H}. 
In fact, the heuristic function defined in~\eqref{eq: H} is not admissible since sometimes it is not the under-estimation of the actual cost-to-goal. However, with the help of~\eqref{eq: H}, we are able to search for trajectories in higher dimensional space in a much faster speed since it tends to search the neighboring regions of the given trajectory instead of exploring the whole state space. In order to guarantee optimality with the inadmissible heuristic in~\eqref{eq: H}, we can combine it with the consistent LQMT heuristic~\cite{liu_iros_2017}, and use multi-heuristic A*~\cite{aine2016multi}.

\begin{figure}[htp]
  \centering
  \subfigure[$\Phi^2: T^2 = 35s, J^2 = 3.0, t^2 = 11ms$]{\includegraphics[width=0.9\columnwidth, trim=0 0 0 0, clip=true]{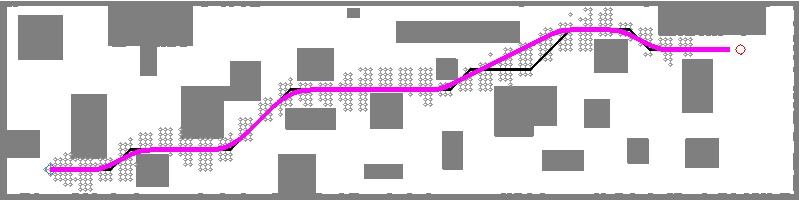}\label{fig: acc}}
  \subfigure[$\Phi^3: T^3 = 36s, J^3 = 4.25, t^3 = 98ms$]{\includegraphics[width=0.9\columnwidth, trim=0 0 0 0, clip=true]{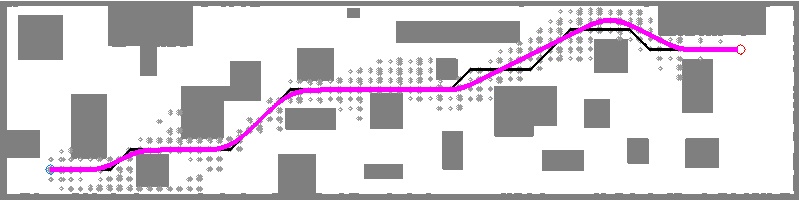}\label{fig: jrk}}
   \caption{Trajectories (magenta) planned using $\Phi^1$ as the heuristic. \label{fig: trajs_p}}
\end{figure}

The results of applying~\eqref{eq: H} for the same planning tasks in Fig.~\ref{fig: trajs} are given in Fig.~\ref{fig: trajs_p}, in which $\Phi^1$ is used as heuristic to plan for both trajectory $\Phi^2$ and $\Phi^3$. Comparing Fig.~\ref{fig: trajs_p} to~\ref{fig: trajs}, the total cost of control effort and execution time, namely $J^q + \rho T^q$, of the new trajectories $\Phi^q$ in Fig.~\ref{fig: trajs_p} are greater than the optimal trajectories in Fig.~\ref{fig: trajs}, but the computation time $t^q$ are much less. 


\section{Evaluation}\label{sec: ana}
\subsection{2-D Planning}
2-D planning is efficient and useful in 2.5-D environments where the obstacles are vertical to the floor. We start by showing 2-D planning tasks of flying though gaps with different widths. In Fig.~\ref{fig: walls} shows how planned trajectories $\Phi^3$ using jerk as a control input vary as the gap in a wall is shrinking (left wall moves closer to the right wall from (a) to (f)). Accordingly, the angle of the desired roll at the gap $\phi_{gap}$ increases. Assume the robot has radius $r=0.35$m, height $h = 0.1$m, and the maximum acceleration in each axis is $a_{max} = g$.  Denoting the roll along trajectory as $\phi$, according to~\eqref{eq: force} and~\eqref{eq: rotation}, we have
\begin{equation}\label{eq:phi}
-\arctan{\frac{a_{max}}{g}} \leq \phi \leq \arctan{\frac{a_{max}}{g}}
\end{equation}
since the desired acceleration in $z$-axis is zero. In other words, the smallest gap that the robot can pass through using 2-D planning is approximately equal to $2r\cos{\theta}$ (which is approximately equal to $0.525$m).

\subsection{3-D Planning}
\begin{figure}[htp]
  \centering
  \subfigure[$\phi_{gap} = 0^\circ$]{\includegraphics[width=0.3\columnwidth, trim=0 1cm 0 1cm, clip=true]{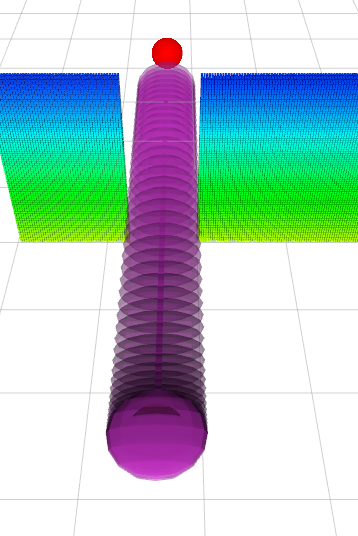}}
  \subfigure[$\phi_{gap} = 27^\circ$]{\includegraphics[width=0.3\columnwidth, trim=0 1cm 0 1cm, clip=true]{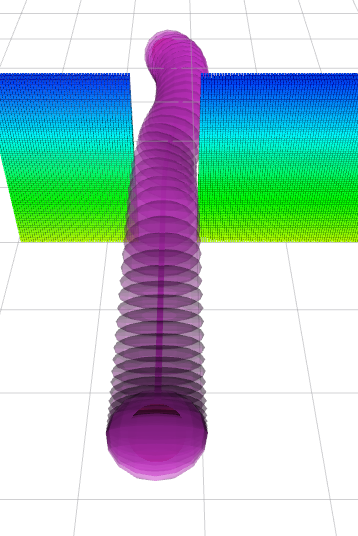}}
  \subfigure[$\phi_{gap} = 45^\circ$]{\includegraphics[width=0.3\columnwidth, trim=0 1cm 0 1cm, clip=true]{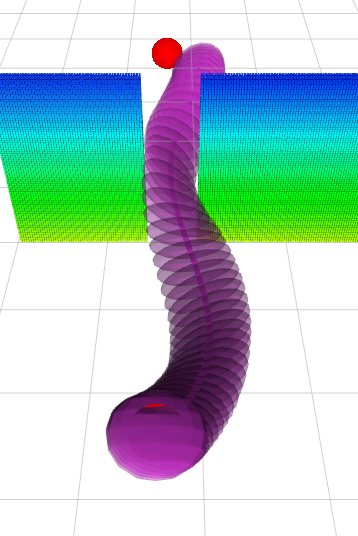}}   
  \subfigure[$\phi_{gap} = 46^\circ$]{\includegraphics[width=0.3\columnwidth, trim=0 1cm 0 1cm, clip=true]{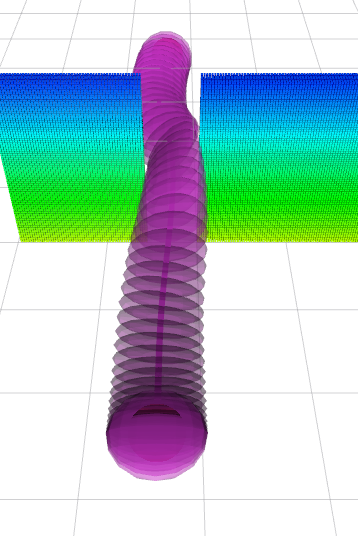}}
  \subfigure[$\phi_{gap} = 73^\circ$]{\includegraphics[width=0.3\columnwidth, trim=0 1cm 0 1cm, clip=true]{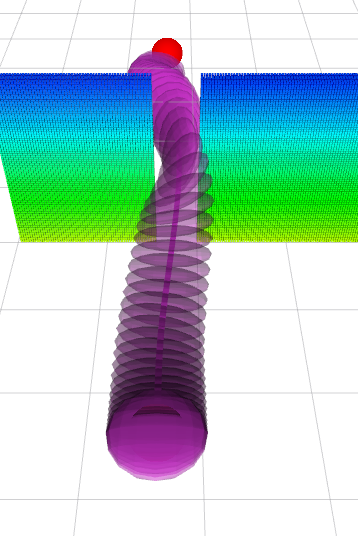}}
   \subfigure[$\phi_{gap} = 90^\circ$]{\includegraphics[width=0.3\columnwidth, trim=0 1cm 0 1cm, clip=true]{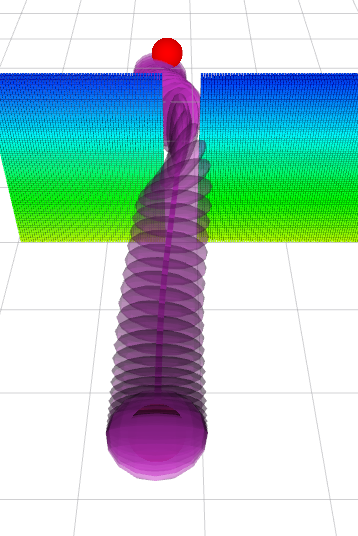}}
   \caption{Trajectories through gaps with different widths: $0.75, 0.65, 0.55$m from (a) to (c) and $0.55, 0.45, 0.35$m from (d) to (f). $\phi_{gap}$ indicates the maximum roll at the gap. Red dots show the start and goal. The top 3 figures show the 2-D planning results, and the bottom 3 figures show the 3-D planning results. \label{fig: walls}}
\end{figure}

By adding control in the $z$-axis, we are able to plan in 3-D space and relax the constraint in~\eqref{eq:phi} as follows:
\begin{equation}\label{eq:phi2}
-\arctan{\frac{a_{max}}{g-a_{max}}} \leq \phi \leq \arctan{\frac{a_{max}}{g-a_{max}}}
\end{equation}
When $a_{max} \geq g$, $\phi \in (-\frac{\pi}{2}, \frac{\pi}{2}]$ can be arbitrary. Letting $a_{max} = g$, we are able to reduce the gap width even more as shown in Fig.~\ref{fig: walls}.

\begin{figure}[htp]
  \centering
  \subfigure[$\phi_{win} = 30^\circ$]{\includegraphics[width=0.3\columnwidth, trim=0 0cm 0 1cm, clip=true]{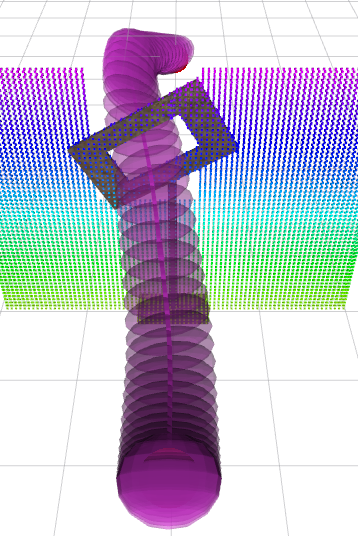}\label{fig: window30}}
  \subfigure[$\phi_{win} = 45^\circ$]{\includegraphics[width=0.3\columnwidth, trim=0 0cm 0 1cm, clip=true]{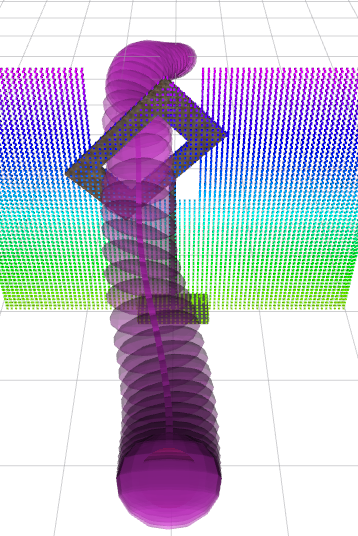}\label{fig: window45}}
  \subfigure[$\phi_{win} = 60^\circ$]{\includegraphics[width=0.3\columnwidth, trim=0 0cm 0 1cm, clip=true]{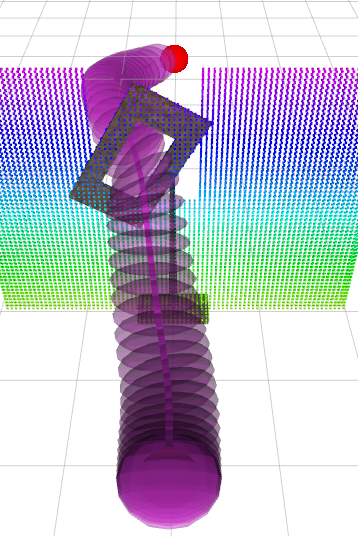}\label{fig: window60}}
   \caption{Trajectories generated through a rectangular hole of size $0.4\times 0.8$m oriented at different angles. A robot with radius $r=0.35$m needs to fly through the hole with certain non-zero roll and pitch angles. The colored dots represent walls in the map that invalidate trajectories that go around the window. \label{fig: window}}
\end{figure}

Another example of 3-D planning using a window with a rectangular hole in the middle is considered. By modifying the angle of the window's inclination $\phi_{win}$, we are able to verify the planner's capability to generate agile trajectories as shown in Fig.~\ref{fig: window}.

\subsection{Parameters}
There are a few parameters that significantly affect the planning performance including computation time, continuity and dynamics constraints. In this section, we analyze these relationships and provide guidance on how to set the parameters in our planner. In the above examples of 2-D and 3-D planning, we used the following settings:

\begin{center}
  \begin{tabular}{  c | c | c | c | c | c }
    \hline
    $\rho$ & $\tau$ & $v_{max}$ & $a_{max}$& $u_{max}$ & $du$  \\ \hline
    $10000$ & $0.2$s & $7\text{m}/\text{s}$ & $10 \text{m}/\text{s}^2$& $50\text{m}/\text{s}^3$ & $12.5\text{m}/\text{s}^3$ \\ 
    \hline
  \end{tabular}
\end{center}

As described in~\cite{liu_iros_2017}, a larger $\rho$ results in faster trajectories. The scale of $\rho$ should be comparable to the scale of the associated control effort. Here we use $\rho \approx 4u_{max}^2$. The motion primitive duration $\tau$ should not be too small or too large. For moderate flight speeds, we find $\tau = 0.2$s to be a reasonable choice. A small $\tau$ makes the graph dense and requires more explorations to reach the goal, while a large $\tau$ may easily result in searching failure since the graph may be too sparse to cover the feasible region. The discretization in the control space $\mathcal{U}_M$ also affects the density of the graph as shown in Fig.~\ref{fig: graph}. Its effect is similar to $\tau$ -- finer discretization in $\mathcal{U}_M$ leads to a slower but more complete search and smoother trajectories and vice versa.

\section{Experiments}
\label{sec: exp}

\subsection{Simulation Results}
\begin{figure}[htp]
  \centering
  \subfigure[Office environment]{\includegraphics[width=0.45\columnwidth, trim=0 0 0 0, clip=true]{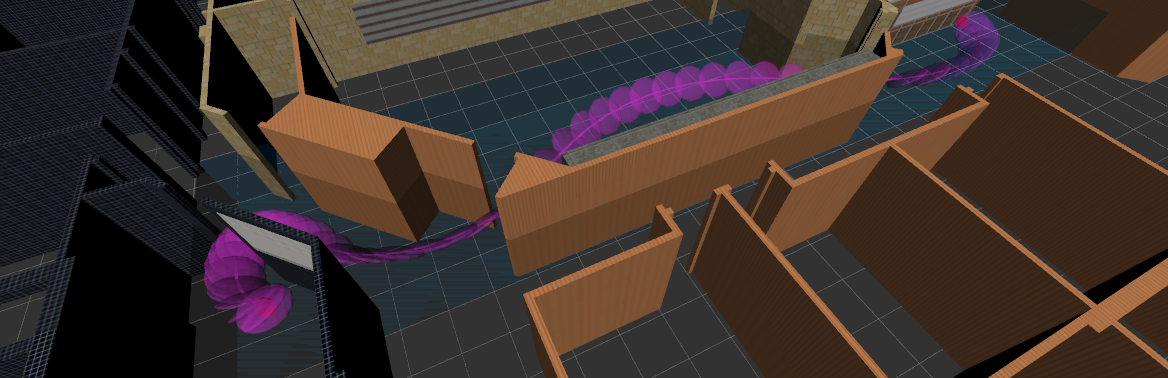}}
  \subfigure[Unstructured environment]{\includegraphics[width=0.45\columnwidth, trim=0 0 0 0, clip=true]{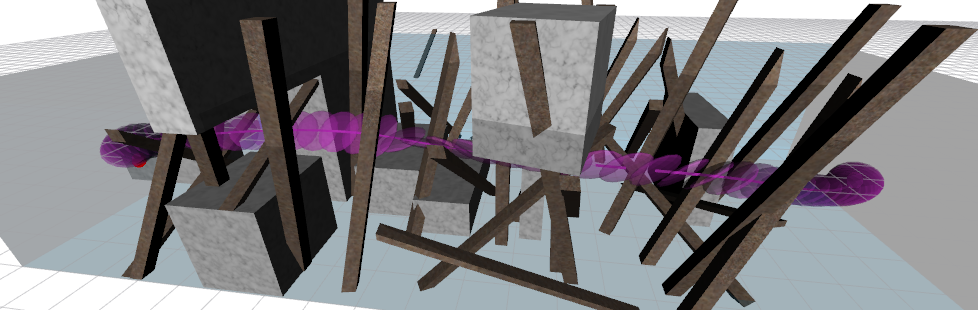}}
   \caption{Generated trajectories in two different environments. The robot radius is $r= 0.5$m, making its diameter much larger than the door width in (a). If the obstacles in these environments are inflated by $r$, no feasible paths exist. \label{fig: sims}}
\end{figure}

The proposed planner is used to generate trajectories in complicated environments as shown in Fig.~\ref{fig: sims}. A geometric model of the environment is converted into a point cloud and used to construct an obstacle \textit{KD-tree}.


\begin{figure}[htp]
  \centering
  \subfigure{\includegraphics[width=0.49\columnwidth, trim=0 0 0 0, clip=true]{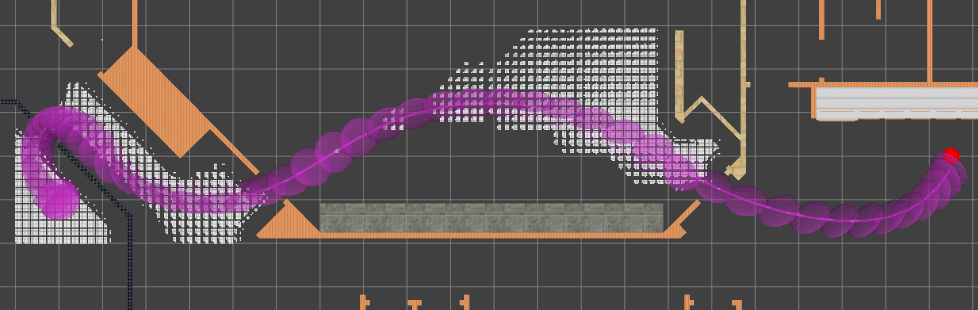}}
  \subfigure{\includegraphics[width=0.49\columnwidth, trim=0 0 0 0, clip=true]{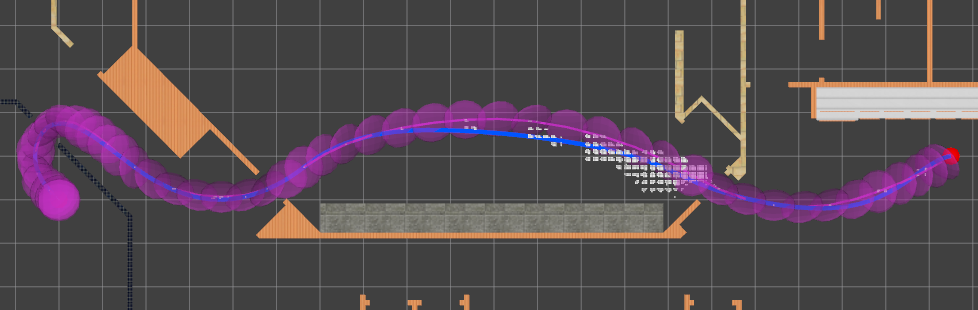}}
  \subfigure{\includegraphics[width=0.49\columnwidth, trim=0 0 0 0, clip=true]{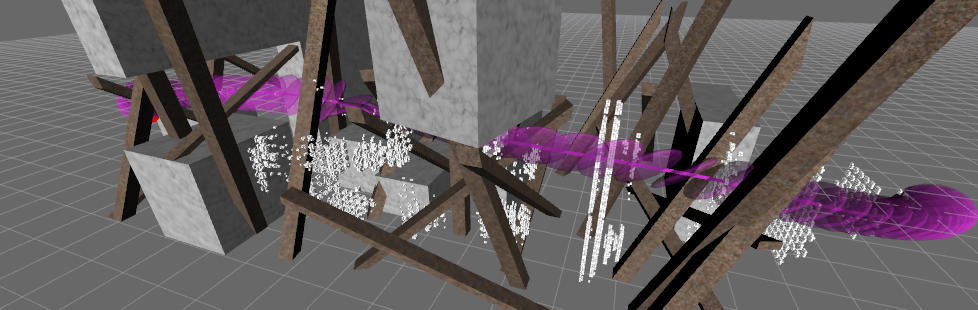}}
  \subfigure{\includegraphics[width=0.49\columnwidth, trim=0 0 0 0, clip=true]{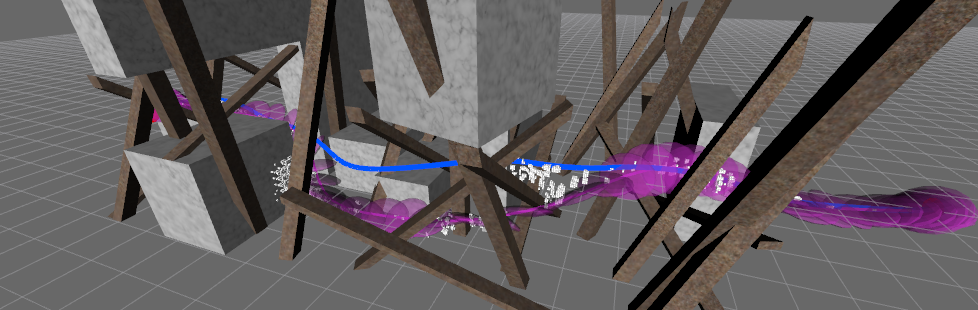}}
   \caption{Comparison between the optimal method (left) and refinement (right). The prior trajectory $\Phi^2$ is plotted in blue, while the white dots indicate explored states. It is clear that the refinement explores fewer irrelevant regions but the generated trajectory is suboptimal.\label{fig: prior}}
\end{figure}

In general, the computation time for finding the optimal trajectories in Fig.~\ref{fig: sims} is slow (Table~\ref{tab: times}). As proposed in Section~\ref{sec: refine}, we plan trajectories $\Phi^2$ using acceleration control at first, based on which we plan the trajectory $\Phi^3_*$ using jerk control. As shown in Table~\ref{tab: times}, the computation time for hierarchical planning is much less than that for planning in the original 9 dimensional space with jerk input. We can also see in Fig.~\ref{fig: prior} that the refinement process tends to explore fewer states. As expected, the refined trajectory $\Phi^3_*$ has a higher cost compared to the optimal trajectory $\Phi^3$.
\begin{table}[H]
\begin{center}
\caption{Evaluation of Trajectory Generation: $t$ refers to the computation time of obtaining a trajectory, $J$ is the total control (jerk) effort and $T$ is the total execution time associated with the trajectory. }\label{tab: times}
  \begin{tabular}{ c || c | c | c | c | c | c }
  \hline
    &  \multicolumn{3}{c|}{Office} &   \multicolumn{3}{c}{Unstructured 3-D}\\
     & $t$(s) & $J(\times 10^3)$ & $T$(s) & $t$(s) & $J(\times 10^3)$ & $T$(s)  \\\hline
    $\Phi^3$ & $89.42$ & $8.9$ & $4.6$ & $129.58$ & $5.6$ & $3.0$ \\  
    $\Phi^2$ & $9.34$ & $0$ & $4.4$ & $21.64$ & 0 & $3.6$ \\  
$\Phi^3_*$ & $2.03$ & $11.1$ & $5.0$ & $24.02$ & $15.1$ & $4.8$ \\      
    \hline
  \end{tabular}
\end{center}
\end{table}

\subsection{Real World Experiments}
\begin{figure}[htp]
  \centering
  \subfigure{\includegraphics[width=0.32\columnwidth, trim=0 0 0 0, clip=true]{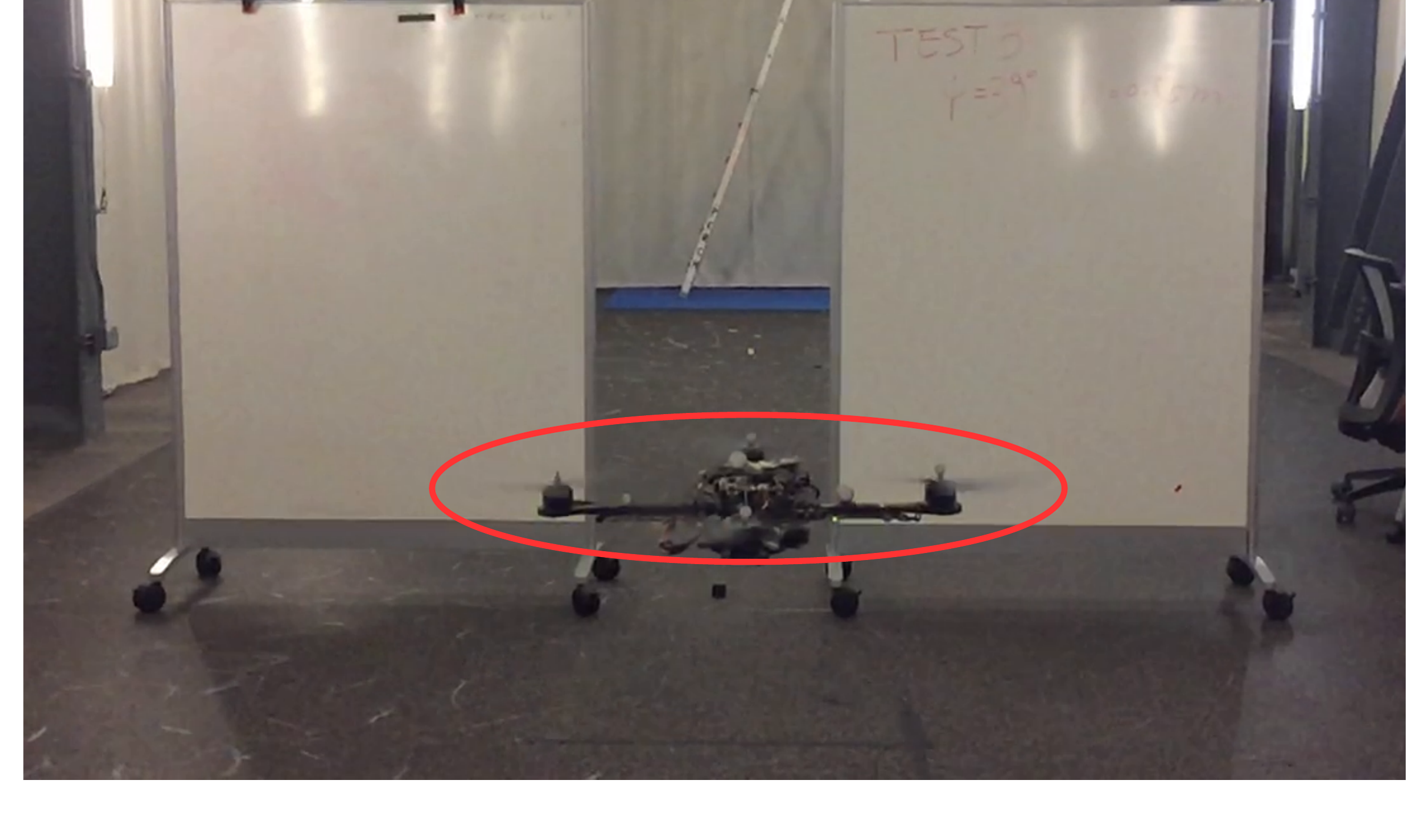}}
  \subfigure{\includegraphics[width=0.32\columnwidth, trim=0 0 0 0, clip=true]{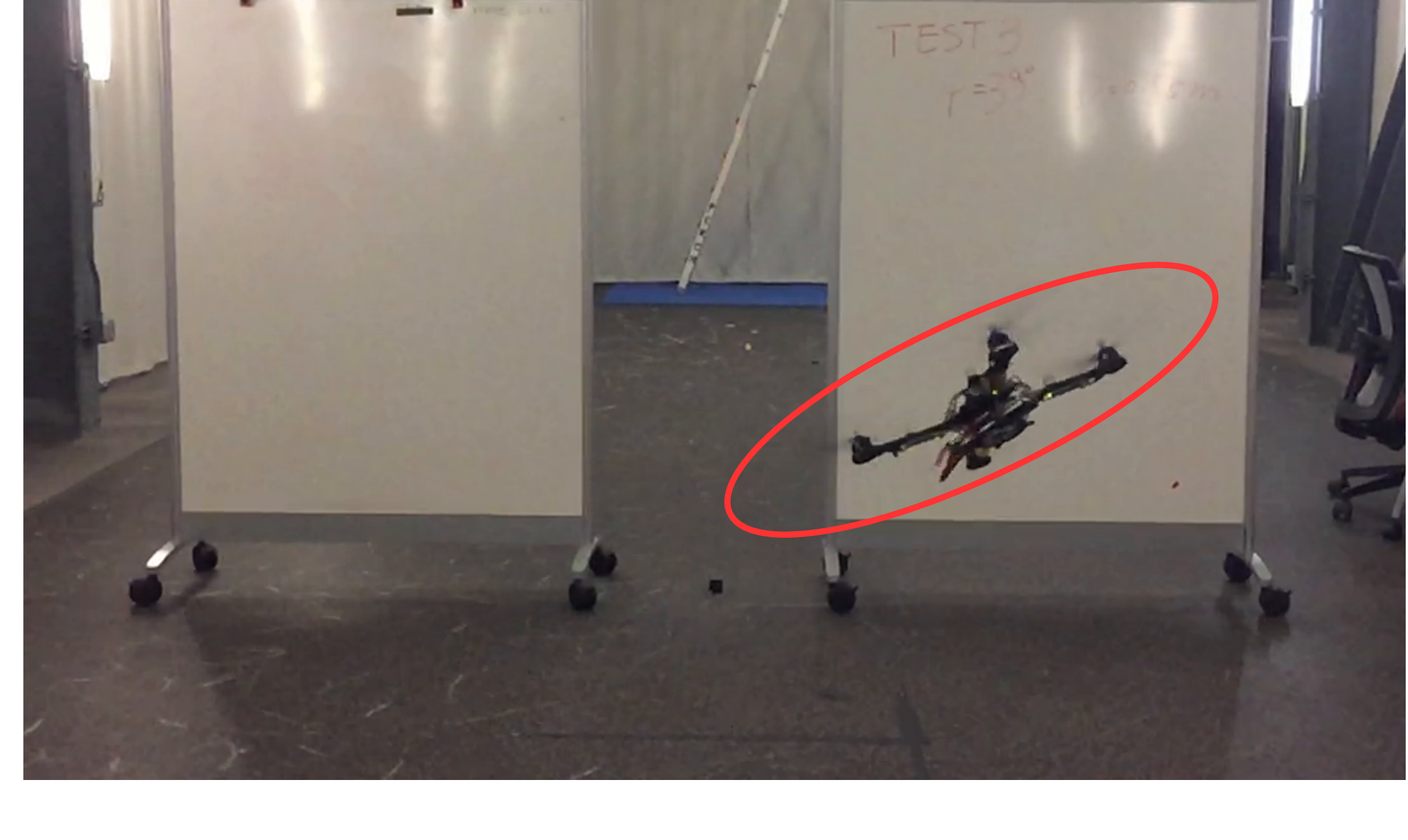}}
  \subfigure{\includegraphics[width=0.32\columnwidth, trim=0 0 0 0, clip=true]{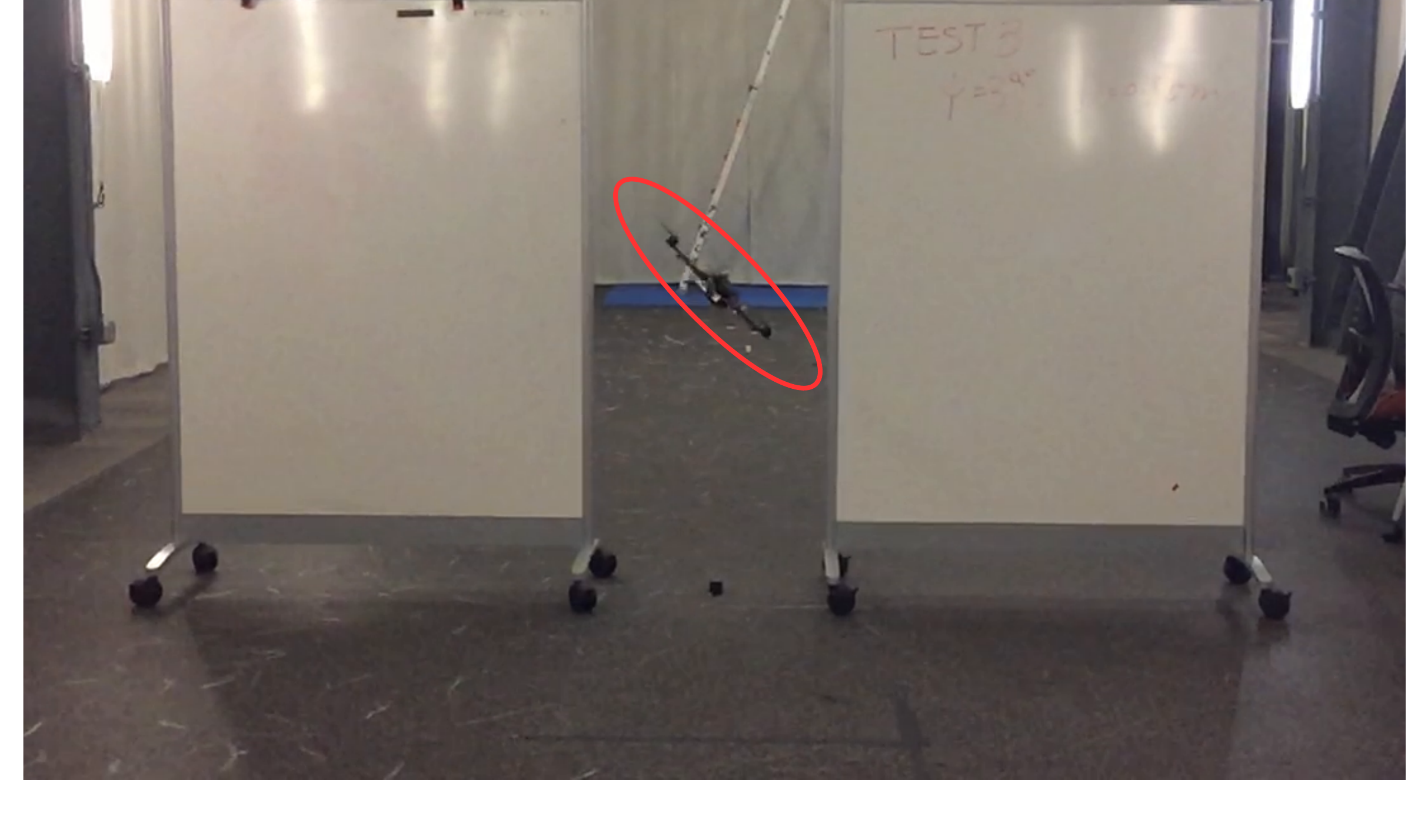}}
  \subfigure{\includegraphics[width=0.32\columnwidth, trim=0 0 0 0, clip=true]{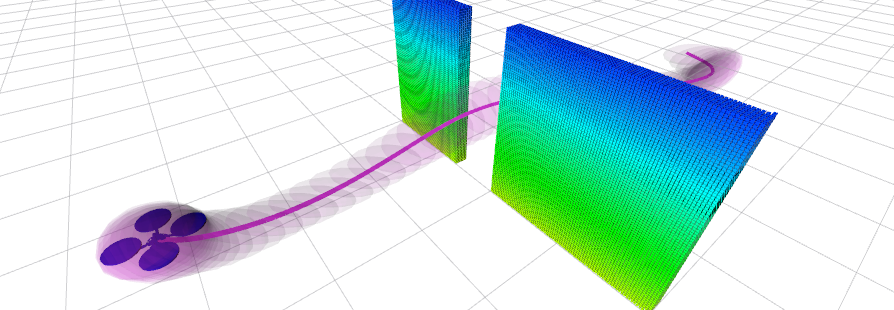}}
  \subfigure{\includegraphics[width=0.32\columnwidth, trim=0 0 0 0, clip=true]{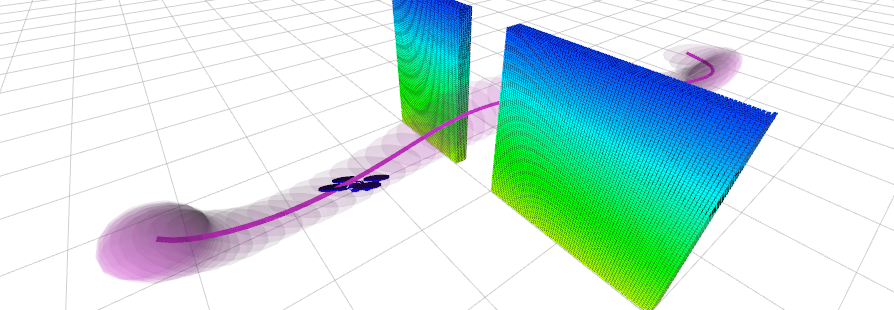}}
  \subfigure{\includegraphics[width=0.32\columnwidth, trim=0 0 0 0, clip=true]{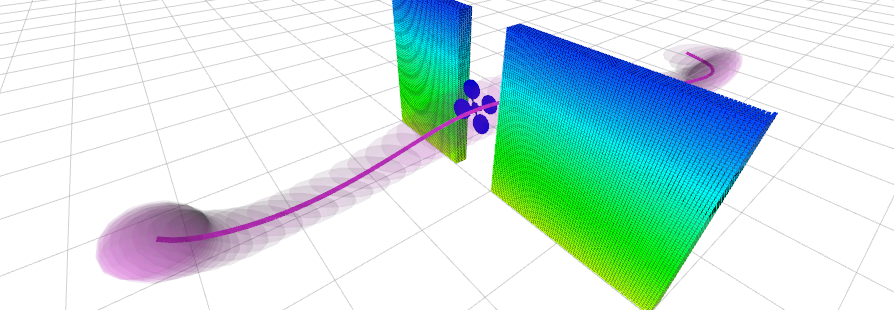}}
   \caption{Quadrotor tracks the planned trajectory to fly through a narrow gap. Top figures are the snapshots of the video, bottom figures are corresponding visualizations in ROS. Maximum roll angle at the gap is $40^\circ$ as drawn in the top right figure.  \label{fig: exp}}
\end{figure}

 The experiments is aiming to demonstrate the feasibility of planned  aggressive trajectories with a real robot. We use AscTec Hummingbird as our quadrotor platform, we also use VICON motion capture system to localize the quadrotor and the obstacle map is obtained by depth sensor in advance to generate trajectories. Using the robust feedback control~\cite{lee2010geometric}, the robot is able to avoid hitting obstacles by following the planned trajectory. Fig.~\ref{fig: exp} shows several snapshots of the flight where the quadrotor needs to roll at $\phi = 40^\circ$ in order to pass through the gap without hitting the white board. 

\begin{figure}[htp]
  \centering
  \subfigure{\includegraphics[width=0.9\columnwidth, trim=0 0 0 0, clip=true]{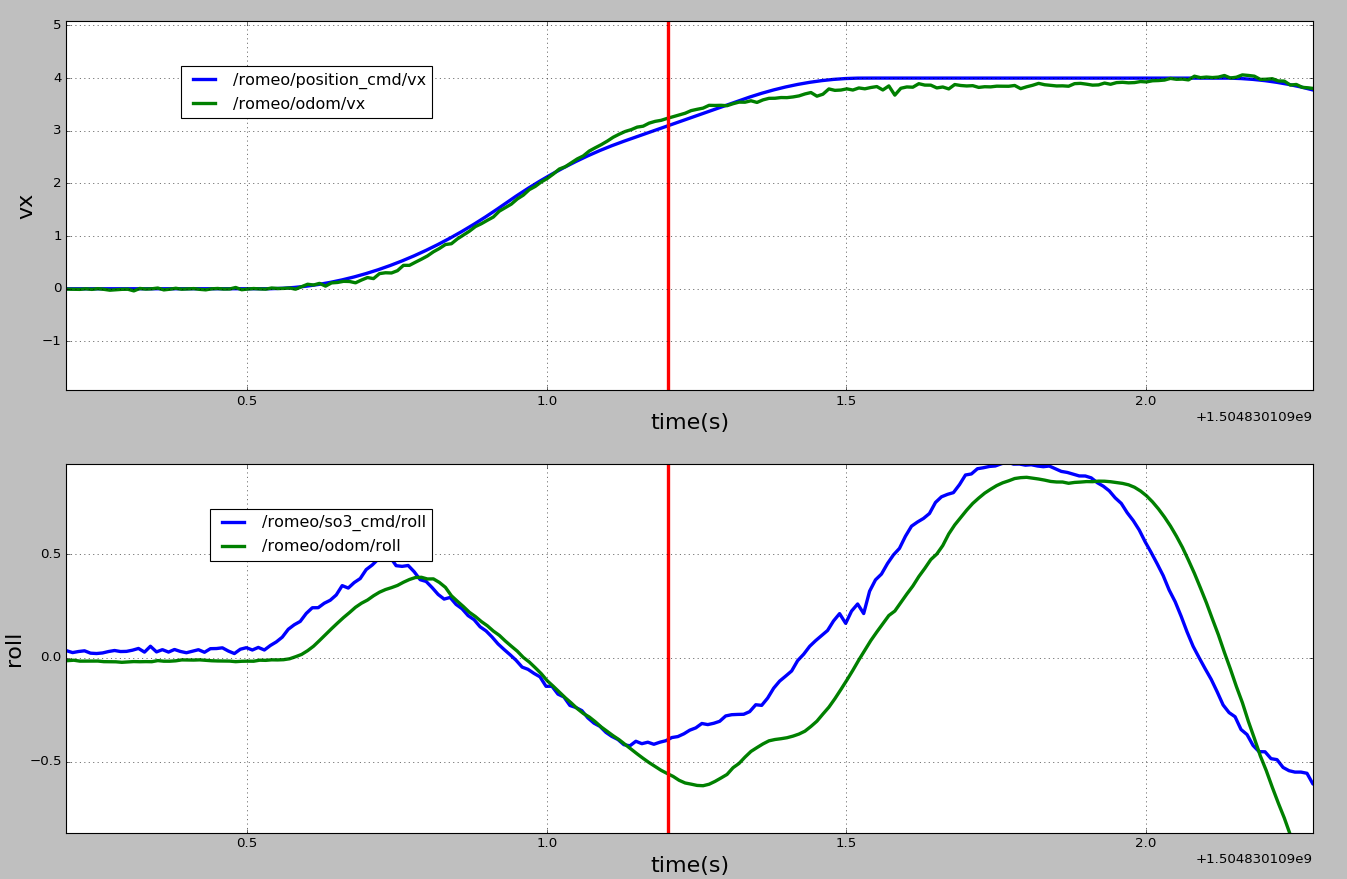}}
     \caption{Plots of control errors, the blue curve is the command value while the green curve shows the actual robot state. Top figure shows $v_x - t$, bottom figure shows $\phi - t$. The red verticle line indicates the time when the robot pass through the gap.  \label{fig: exp_plot}}
\end{figure}

The control errors in velocity and roll are plotted in Fig.~\ref{fig: exp_plot}. The commanded roll includes the feedback attitude errors such that it is not as smooth as the desired  roll from the planned trajectory. The robot is able to track velocity properly up to $4m/s$, but clearly there exists lag in the attitude control. This is because the actual robot is not able to achieve specified angular velocity instantly due to the dynamics. A more accurate model for the quadrotor is to use snap as the control input instead of the jerk. The trajectory planned using the snap as the control input is straightforward to solve following the same pipeline as proposed in this paper, which has also been implemented in our open-sourced planner. 

\section{Conclusion}
\label{sec: con}
In this work, we extend our previous motion-primitive-based planning algorithm~\cite{liu_iros_2017} to enable aggressive flight with attitude constraints in cluttered environments for an under-actuated quadrotor system. We also presented a hierarchical refinement process that uses prior lower-dimensional trajectories as heuristics to accelerate planning in higher dimensions. Our planner is the first to plan dynamic trajectories in cluttered environments in SE(3) while guaranteeing safety, trajectory smoothness, and optimality. We believe that in future work, it is possible to integrate the planner with onboard sensing, state estimation, and feedback control to obtain a fully autonomous quadrotor system that is able to fly aggressively but safely in unknown cluttered environments.


\appendices
\section{}
\label{app:a}
\subsection*{Linear Quadratic Minimum Time for Jerk Control}
The heuristic function $H(\mathbf{s}, \mathbf{s}_g)$ for graph search is an under-estimation of actual cost from the current state $\mathbf{s}$ to the goal state $\mathbf{s}_g$ by relaxing the dynamics and obstacles constraints. We try to find a state-to-state optimal trajectory of Problem~\ref{prob:3}, whose cost serves as the cost-to-go heuristic $H$. The explicit solution for the optimal cost for velocity, acceleration control has been shown in~\cite{liu_iros_2017}. Here we show the explicit solution for jerk control.
\begin{problem}\label{prob:3}
  Given a current state $\mathbf{s}$, the goal state $\mathbf{s}_g$, find the optimal trajectory according to the cost function
  \begin{equation}
    \label{eq:problem3}
    \begin{gathered}
      \min_{j, T} \; \int_{0}^{T}  j^2dt + \rho T
    \end{gathered}
  \end{equation}
\end{problem}
Assume the initial state is given as  $\mathbf{s} = [p_0, v_0, a_0]^\T$, the formulation of the optimal trajectory for~\eqref{eq:problem3} is given from the Pontryagin's minimum principle~\cite{mueller2015} as
\begin{align}
p = \frac{d_5}{120}t^5+\frac{d_4}{24}t^4+\frac{d_3}{6}t^3+\frac{a_0}{2}t^2+v_0t+p_0
\end{align}

The coefficients $[d_5, d_4, d_3]$ are defined in~\cite{mueller2015} by $\mathbf{s}, \mathbf{s}_g$ and $T$. 
As a result, the total cost of~\eqref{eq:problem3} can be written as a function of time $T$ as
\begin{align}\label{eq: C}
\mathcal{C}(T) =\;&  \int_{0}^{T}  (\frac{d_5}{2}t^2 + d_4t + d_3)^2dt + \rho T \nonumber\\
 =\; &\frac{d_5^2}{20}T^5+\frac{d_4d_5}{4}T^4+(\frac{d_4^2}{3}+\frac{d_3d_5}{3})T^3\\
&+d_3d_4T^2+d_3^2T + \rho T \nonumber
\end{align}

The minimum of $\mathcal{C}(T)$ can be derived by taking the derivative with respect to $T$ and finding the root $T^*$ of
\begin{equation}\label{eq: C'}
\frac{d\mathcal{C}}{dT} = c_0+\hdots+c_6T^{-6} =  0,\;  T \in [0, \infty)
\end{equation}
Therefore, $H(\mathbf{s}, \mathbf{s}_g) = \mathcal{C}(T^*)$. The coefficients in~\eqref{eq: C'} are derived as follows:\\
(1) Fully Defined $\mathbf{s}_g =  [p_1, v_1, a_1]^\T$
\begin{align}
c_0 &= \rho,\;c_1 = 0,\;c_2= -9a_0^2+6a_0a_1-9a_1^2,\nonumber\\
c_3&=-144a_0v_0-96a_0v_1+96a_1v_0+144a_1v_1,\\
c_4&=360(a_0-a_1)(p_0-p_1)-576v_0^2-1008v_0v_1-576v_1^2,\nonumber\\
c_5&= 2880(v_0+v_1)(p_0-p_1),\nonumber\\
c_6& = -3600(p_0-p_1)^2.\nonumber
\end{align}
(2) Partially Defined $\mathbf{s}_g = [p_1, v_1]^\T$
\begin{align}
c_0 &= \rho,\;c_1 = 0,\;c_2= -8a_0^2,\nonumber\\
c_3&=-112a_0v_0-48a_0v_1,\\
c_4&=240a_0(p_0-p_1)-384v_0^2-432v_0v_1-144v_1^2,\nonumber\\
c_5&= (1600v_0+960v_1)(p_0-p_1),\nonumber\\
c_6& = -1600(p_0-p_1)^2.\nonumber
\end{align}
(3) Partially Defined $\mathbf{s}_g = [p_1]^\T$
\begin{align}
c_0 &= \rho,\;c_1 = 0,\;c_2= -5a_0^2,\nonumber\\
c_3&=-40a_0v_0,\\
c_4&=60a_0(p_0-p_1)-60v_0^2,\quad\quad\quad\nonumber\\
c_5&= 160v_0(p_0-p_1),\nonumber\\
c_6& = -100(p_0-p_1)^2.\nonumber
\end{align}


\bibliographystyle{bib/IEEEtran}
\bibliography{bib/references}

\end{document}